\documentclass{article}

\usepackage{microtype}
\usepackage{graphicx}
\usepackage{subfigure}
\usepackage{booktabs} 

\usepackage{hyperref}

\usepackage[accepted]{icml2024}

\usepackage{amsmath}
\usepackage{amssymb}
\usepackage{mathtools}
\usepackage{amsthm}

\usepackage[capitalize,noabbrev]{cleveref}

\theoremstyle{plain}

\theoremstyle{definition}

\theoremstyle{remark}

\usepackage[textsize=tiny]{todonotes}

\usepackage{subfiles}

\icmltitlerunning{Iterated Energy-based Flow Matching for Sampling from Boltzmann Densities}

\begin{document}

\twocolumn[
\icmltitle{Iterated Energy-based Flow Matching for Sampling from Boltzmann Densities}



\icmlsetsymbol{equal}{*}

\begin{icmlauthorlist}
\icmlauthor{Dongyeop Woo}{sch}
\icmlauthor{Sungsoo Ahn}{sch}


\end{icmlauthorlist}

\icmlaffiliation{sch}{Pohang University of Science and Technology}

\icmlcorrespondingauthor{Dongyeop Woo}{dongyeop.woo@postech.ac.kr}
\icmlcorrespondingauthor{Sungsoo Ahn}{sungsoo.ahn@postech.ac.kr}

\icmlkeywords{Machine Learning, ICML}

\vskip 0.3in
]



\printAffiliationsAndNotice{} 

\begin{abstract}
In this work, we consider the problem of training a generator from evaluations of energy functions or unnormalized densities. This is a fundamental problem in probabilistic inference, which is crucial for scientific applications such as learning the 3D coordinate distribution of a molecule. To solve this problem, we propose iterated energy-based flow matching (iEFM), the first off-policy approach to train continuous normalizing flow~(CNF) models from unnormalized densities. We introduce the simulation-free energy-based flow matching objective, which trains the model to predict the Monte Carlo estimation of the marginal vector field constructed from known energy functions. Our framework is general and can be extended to variance-exploding (VE) and optimal transport (OT) conditional probability paths. We evaluate iEFM on a two-dimensional Gaussian mixture model (GMM) and an eight-dimensional four-particle double-well potential (DW-4) energy function. Our results demonstrate that iEFM outperforms existing methods, showcasing its potential for efficient and scalable probabilistic modeling in complex high-dimensional systems.
\end{abstract}

\section{Introduction}
In this work, we are interested in training a neural network to sample from the distribution $q(x)\propto \exp(-\mathcal{E}(x))$ where $\mathcal{E}(x)$ is a known energy function. A significant example of this problem includes sampling from the Boltzmann distribution of a molecule, where the 3D atomic coordinates $x$ are distributed according to the physical energy of the molecular system. An accurate sampler for this problem is crucial for many scientific applications, e.g., inferring the probability that a protein will be folded at a given temperature for drug discovery \citep{noe2019boltzmann}.

However, the recently successful deep generative models are not applicable to this problem. Unlike the conventional settings with access to samples from the target distribution, creating the dataset is expensive for these problems. One could resort to simulating the actual molecular dynamics or leverage Monte Carlo (MC) techniques such as annealed importance sampling \citep{neal2001annealed} or sequential Monte Carlo~\citep{del2006sequential}. However, such sampling algorithms are computationally expensive and do not easily scale to high dimensions.

To solve this problem, researchers have developed frameworks to directly train generative models using energy function evaluations. A simple approach is to train the generative models to minimize the reverse Kullback-Leibler (KL) divergence estimated by samples from the generator weighted by the unnormalized densities \citep{wirnsberger2022normalizing}. However, this approach suffers from the mode-seeking behaviour that leads to a generator that only captures a particular mode of the system. Alternatively, researchers have considered minimizing the combination of the forward and the reverse KL divergence \citep{noe2019boltzmann} or the $\alpha$-divergence \citep{midgley2023flow}. These works depended on the normalizing flow architecture which allows explicit evaluation of the data likelihood. 

Without using the normalizing flow models, researchers have also considered training diffusion-based samplers, e.g., path integral sampler \citep[PIS]{zhang2021path}, time-reversed diffusion sampler \citep[DIS]{berner2024an}, denoising diffusion sampler \citep[DDS]{vargas2022denoising}. However, they mostly require simulation of the diffusion-based models to generate trajectories to train on, which may be expensive to collect. In addition, their objective is on-policy, and they cannot reuse samples collected from previous versions of the generative model without importance sampling.

Recently, iterated denoising energy matching \citep[iDEM]{akhound2024iterated} has been proposed as a simulation-free and off-policy approach to train the generators. They propose to train the denoising diffusion model to generate the samples. The generator is trained on Monte Carlo estimation of the score function. It consists of iteratively collecting samples and estimating the score function. Continuous generative flow network \citep[GFlowNet]{bengio2021flow, lahlou2023theory, sendera2024diffusion} also serves as a framework to train the diffusion models using energy functions in a simulation-free and off-policy manner.

In this work, we consider the first simulation-free and off-policy framework to learn the Boltzmann distribution using a continuous normalizing flow (CNF), an alternative to the current diffusion-based models. Unlike the diffusion-based models, CNFs deterministically generate the data using vector fields. They are promising in their flexibility to be generalized across a variety of probability paths \citep{lipman2022flow}, with variations that are faster and easier to train compared to the diffusion-based models \citep{tong2024improving}.


In particular, our key idea is to estimate the flow matching objective using a newly derived Monte Carlo estimation of the data-generating vector field. Our algorithm is structured similarly to the iDEM algorithm. We employ a simulation-free and off-policy objective which involves estimation of the data-generating vector field to train the CNF. We also use a replay buffer to reuse the samples and improve sample-efficiency for training. We consider the variance exploding and conditional optimal transport probability paths to implement our idea.


We evaluate our algorithm for the two-dimensional Gaussian mixture model (GMM) and eight-dimensional four-particle double-well potential (DW-4) energy functions. Our iEFM is shown to outperform the existing works, including iDEM, demonstrating the promise of our work.
\section{Iterated Energy-based Flow Matching}

\subsection{Flow Matching Objective}
In this work, we are interested in learning a Boltzmann distribution $\mu(x)$ associated with an energy function $\mathcal{E}: \mathbb{R}^d \rightarrow \mathbb{R}$ defined as follows:
\begin{equation}
     \mu(x) = \frac{\exp(-\mathcal{E}(x))}{Z}, \quad Z = \int_{\mathbb{R}^d} \exp(-\mathcal{E}(x))dx,
\end{equation}
where $Z$ is the intractable normalizing constant. Unlike the conventional settings of training a generative model, our scheme does not require samples from $\mu(x)$, but assumes the ability to evaluate the energy function $\mathcal{E}(x)$.

Our scheme aims to train a continuous normalizing flow~(CNF) to match a random process~$x_{t}$ with marginal density $p_t$ for $t \in [0,1]$ which starts at simple Gaussian prior $p_{0}(x_{0}) = N(x_{0}; 0, \sigma_0^2)$ and ends at the target distribution $p_1(x_{1}) = \mu(x_{1})$. In particular, we let $u_{t}(x; \theta)$ parameterized by a neural network regress the time-dependent vector field $v_t: \mathbb{R}^d \rightarrow \mathbb{R}^d$ with $t \in [0,1]$ that generates the marginal density $p_{t}$. In particular, we consider the following flow matching~(FM) objective \citep{lipman2022flow}:
\begin{equation}\label{eq:fm}
    \mathcal{L}_{\text{FM}}(\theta) = \mathbb{E}_{t\sim [0,1], x\sim p_{t}(x)}[\lVert u_t(x; \theta) - v_t(x) \rVert_2^2],
\end{equation} 
where $t$ is sampled uniformly from $[0,1]$. Consequently, one can sample from the CNF by starting from an initial value $x_0$ from the prior distribution $p_0$ and solving the ordinary differential equation~(ODE) expressed by~$dx = u_t(x;\theta)dt$. We provide a more detailed description of CNF in \cref{appx:cnf}.

\begin{algorithm}[t]
   \caption{Iterated Energy-based Flow Matching}
   \label{alg:iEFM}
\begin{algorithmic}
   \STATE {\bfseries Input:} Network $u_\theta$, noise schedule $\sigma_t^2$, prior $p_0$, replay buffer $\mathcal{B}$, number of data point added to replay buffer $B_1$, batch size $B_2$, and number of MC samples $K$.
   \WHILE{ Outer-Loop }
        \STATE Set $\{x_0^{(b)}\}_{b=1}^{B_1} \sim p_0(x_0)$.
        \STATE Set $\{x_1^{(b)}\}_{b=1}^{B_1} \leftarrow \text{ODEsolve}(\{x_0^{(b)}\}_{b=1}^{B_{1}}, u_\theta, 0, 1)$.
        \STATE Update the buffer $\mathcal{B} = (\mathcal{B} \cup \{x_1^{(b)}\}_{b=1}^{B_1})$.
        \WHILE{ Inner-Loop }
            \STATE Sample $\{x_1^{(b)}\}_{b=1}^{B_2}$ from the buffer $\mathcal{B}$.
            \STATE Sample $t^{(b)}$ uniformly from $[0, 1]$.
            \STATE Sample $x^{(b)} \sim p_{t^{(b)}}(x^{(b)} |x_1^{(b)})$ for $b= 1, \dots, B_2$.
            \STATE Compute the loss function $\mathcal{L}_{\text{EFM}}$ defined as: 
            $$\mathcal{L}_{\text{EFM}} = \sum_{b=1}^{B_{2}} \lVert u_{t^{(b)}}(x^{(b)}, \theta) - U_K(x^{(b)}, t^{(b)}) \rVert^2.$$
            \STATE Update $\theta$ to minimize $\mathcal{L}_{\text{EFM}}$.
        \ENDWHILE
    \ENDWHILE
\end{algorithmic}
\end{algorithm}

\subsection{Iterated Energy-based Flow Matching (iEFM)}
Here, we propose our iterated energy-based flow matching~(iEFM) for simulation-free and off-policy training of CNFs using energy functions of the Boltzmann distribution. To this end, we propose a new estimator for the marginal vector field $v_{t}(x)$ in \cref{eq:fm} and describe an iterative scheme to collect samples and update the CNF models. We provide a complete description of iEFM in \cref{alg:iEFM}. 



\textbf{Energy-based flow matching.} We first propose our new objective, which involves estimation of the target vector field $v_{t}(x)$ that constructs the Boltzmann distribution. Among the possible vector fields, we consider the following form proposed in \citet{lipman2022flow}:
\begin{equation}
    \label{eqn:vt_as_integral}
    v_t(x) = \int {v_t(x|x_1)}\frac{p_t(x|x_1)p_1(x_1)}{p_t(x)}dx_1,
\end{equation}
where $v_t(x|x_1)$ is the conditional vector field that generates the conditional probability path $p_t(x|x_1)$. We note how various conditional vector fields can be used to generate the data distribution, e.g., variance exploding, variance preserving, and optimal transport conditional vector fields. 

Our key idea is to express the target vector field $v_{t}(x)$ using ratio of expectations over a distribution $q(x_1;x, t) \propto p_{t}(x|x_{1})$ and apply Monte Carlo estimation. A detailed derivation is as follows:
\begin{equation}
    \begin{split}
    v_t(x)&= \frac{\int v_t(x|x_1) p_t(x|x_1)p_1(x_1)dx_1}{\int p_t(x|x_1)p(x_1) dx_1} \\
    &= \frac{\int v_t(x|x_1) q(x_1; x, t)p_1(x_1)dx_1}{\int q(x_1; x, t)p(x_1) dx_1} \\
    &= \frac{\mathbb{E}_{x_1 \sim q(x_{1};x, t)}[v_t(x|x_1)p_1(x_1)]}{\mathbb{E}_{x_1 \sim q(x_{1};x, t)}[p(x_1)]}.
    \label{eqn:vt_as_expectation}
\end{split}
\end{equation}
From \cref{eqn:vt_as_expectation}, one can estimate $v_{t}(x)\approx U_{K}(x,t)$:
\begin{equation}
    \begin{split}
    U_K(x, t) &= \frac{\frac{1}{K}\sum_{i=1}^{K} v_t(x|x_1^{(i)})p_1(x_1^{(i)})}{\frac{1}{K}\sum_{i=1}^{K} p_1(x_1^{(i)})} \\
    &= \frac{\sum_{i=1}^{K} v_t(x|x_1^{(i)})p_1(x_1^{(i)})}{\sum_{i=1}^{K} p_1(x_1^{(i)})},
    \end{split}
\end{equation}
where $x_1^{(1)}, \dots, x_1^{(K)}$ are $K$ samples from the auxiliary distribution $q(x_{1} ;x, t)$. Intuitively, this marginal vector field estimator $U_K(x, t)$ can also be understood as a weighted estimate of conditional vector fields:
\begin{equation}
    \begin{split}
    \label{eqn:U_K_definition}
    U_K(x, t) &= \sum_{i=1}^{K} w_i v_t(x|x_1^{(i)}), \\
    w_i &:= \frac{p_1(x_1^{(i)})}{\sum_j p_1(x_1^{(j)})}.
    \end{split}
\end{equation}
Here, each conditional vector field term describes the direction in which the probability mass moves to reach sampled endpoint candidates $x_1^{(i)}$.


Finally, we derive our EFM objective as an approximation of the FM objective in \cref{eq:fm} as follows:
\begin{equation}
    \label{eqn:EFM_definition}
    \mathcal{L}_{\text{EFM}}(\theta) = \mathbb{E}_{t\sim[0,1], x\sim r_t(x)}\lVert u_t(x; \theta) - U_K(x, t) \rVert^2, 
\end{equation}
where $r_t(x)$ is a reference distribution with support~$\mathbb{R}^{d}$. It is noteworthy how the original FM objective was proposed with $r_t(x) = p_{t}(x)$, but the optimal vector field does not change with a different choice of reference distribution. Hence, similar to the iDEM approach \citep{akhound2024iterated}, our algorithm is off-policy. Furthermore, evaluating the EFM objective does not require any simulation of the CNF, hence our EFM objective is also a simulation-free approach.

\textbf{Iterative training with replay buffer.} Since our energy-based flow matching objective is off-policy, we employ an additional replay buffer to store previously used samples. In particular, our iEFM algorithm iterates between two steps (a) storing the samples from the CNF into the replay buffer and (b) training the CNF using the energy-based flow matching objective defined on samples from the replay buffer. 


\textbf{Conditional probability paths.} Computation of the EFM objective requires sampling from the distribution $q(x_{1};x,t) \propto p_{t}(x | x_{1})$ which depends on choice of the conditional probability path. In general, we consider a conditional probability path of the following form:
\begin{equation}
    p_t(x|x_1) = \mathcal{N}(x; \mu_t(x_1), \sigma_t^2\mathbf{I}),
\end{equation}
where $\mu_t: \mathbb{R}^{d}\rightarrow \mathbb{R}^{d}$ is a time-dependent invertible differentiable function, $\sigma_{t} \in \mathbb{R}$ is a time-dependent scalar, and $\mathbf{I} \in \mathbb{R}^{d\times d}$ is an identity matrix. 
For optimal transport (OT) conditional probability paths, we set:
\begin{equation}
    p_{t}(x | x_{1}) = \mathcal{N}(x; tx_1, (1 - (1 - \sigma_{1})t)^2\mathbf{I}),
\end{equation}
where $\sigma_{1}$ is a fixed hyper-parameter.
Then the distribution $q(x_{1}; x, t)\propto p_{t}(x | x_{1})$ corresponds to the following distribution:
\begin{equation}
    \label{eqn:q_of_OT}
    q(x_1; x, t) = \mathcal{N}\bigg(x_1; \frac{x}{t}, \bigg(\frac{\sigma_t^{2}}{t^{2}}\bigg) \mathbf{I}\bigg),
\end{equation}
where $\mathbf{I}$ is the identity matrix. For variance exploding (VE) conditional probability paths, $p_{t}(x | x_{1})$ is defined by the mean $\mu_t(x_1) = x_1$ and standard deviation $\sigma_t$ following a geometric noise schedule, which leads to the distribution $q(x_{1}; x, t)\propto p_{t}(x | x_{1})$ expressed as follows:
\begin{equation}
    \label{eqn:q_of_VE}
    q(x_1; x, t) = \mathcal{N}(x_1; x, \sigma_t^2 I).
\end{equation}

\section{Experiments}

\begin{figure}[t]
    \begin{center}
    \centerline{\includegraphics[width=\columnwidth]{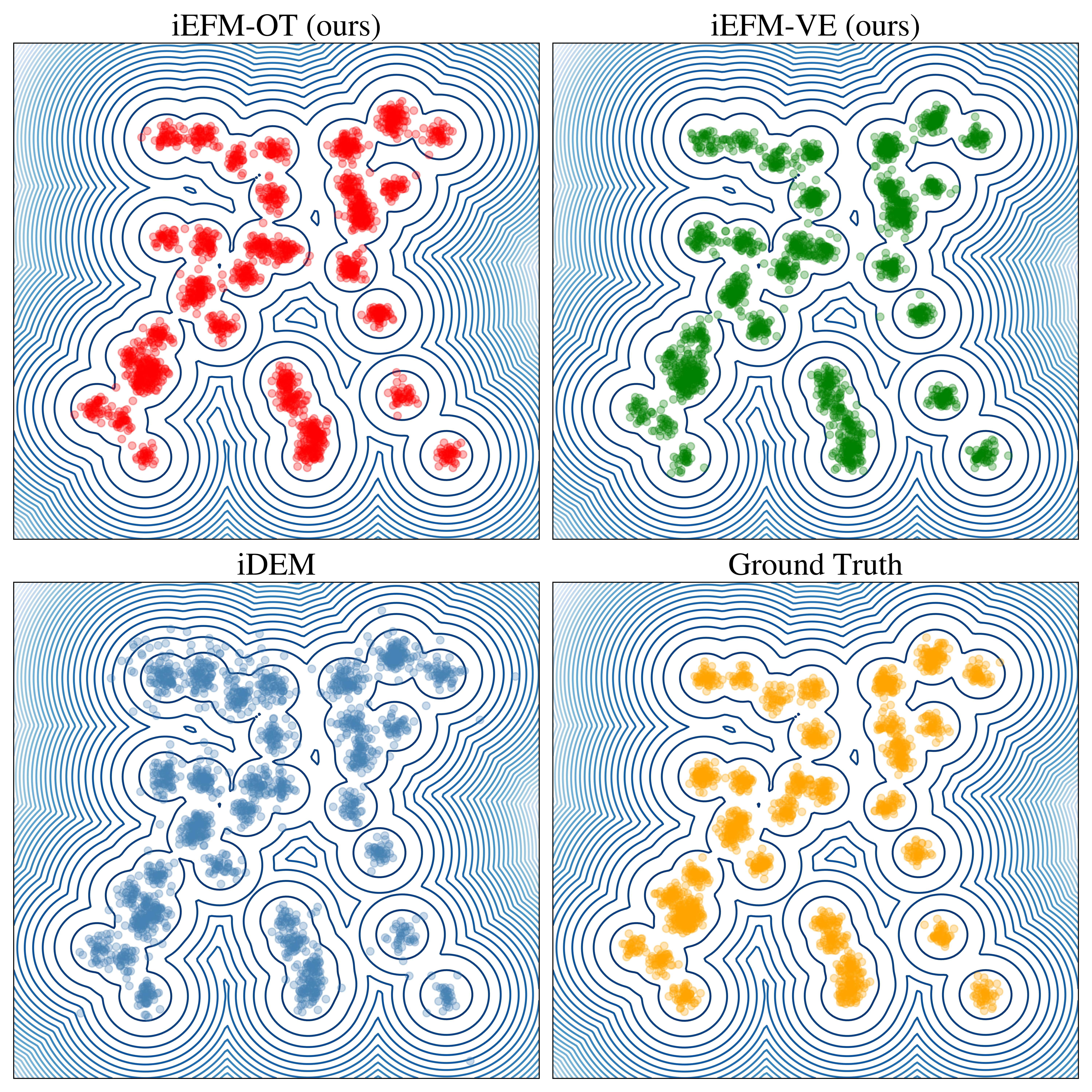}}
    \caption{Contour plot for the energy function of a GMM with 40 modes. Colored points represent samples from each method.}
    \label{fig:gmm_sample}
    \end{center}
    \vskip -0.4in
\end{figure}

\begin{table*}[t]
\caption{Performance comparison for Gaussian mixture model (GMM) and 4-particle double-well (DW-4) energy function. The performance is measured using negative log-likelihood (NLL) and 2-Wasserstein metrics ($\mathcal{W}_2$). We report the results with mean and standard deviation measured over three independent runs. We mark results within the standard deviation of the best number in \textbf{bold}. $^{\dagger}$We compare with the numbers reported by \citet{akhound2024iterated}.}
\label{table:main table}
\vskip 0.15in
\begin{center}
\begin{tabular}{lcccc}
\toprule
Energy $\rightarrow$ & \multicolumn{2}{c}{GMM ($d=2$)} & \multicolumn{2}{c}{DW-4 ($d=8$)} \\
\cmidrule(lr){2-3}\cmidrule(lr){4-5}
Method $\downarrow$ & NLL & $\mathcal{W}_2$ & NLL & $\mathcal{W}_2$ \\
\midrule
FAB$^{\dagger}$ \citep{midgley2023flow}             & 7.14 $\pm$ 0.01 & 12.0 $\pm$ 5.73 & \textbf{7.16 $\pm$ 0.01} & 2.15$\pm$ 0.02 \\
PIS$^{\dagger}$ \citep{zhang2021path}              & 7.72 $\pm$ 0.03 & 7.64 $\pm$ 0.92 & 7.19 $\pm$ 0.01 & 2.13 $\pm$ 0.02 \\
DDS$^{\dagger}$ \citep{vargas2022denoising}              & 7.43 $\pm$ 0.46 & 9.31 $\pm$ 0.82 & 11.3 $\pm$ 1.24 & 2.15$\pm$ 0.04 \\
iDEM$^{\dagger}$ \citep{akhound2024iterated}             & \textbf{6.96 $\pm$ 0.07} & 7.42 $\pm$ 3.44 & \textbf{7.17 $\pm$ 0.00} & 
2.13 $\pm$ 0.04 \\
\midrule
iEFM-VE (ours)   & 7.08 $\pm$ 0.04    & \textbf{4.91 $\pm$ 0.60} & 7.53 $\pm$ 0.00 & 2.21 $\pm$ 0.00 \\
iEFM-OT (ours)   & \textbf{6.92 $\pm$ 0.05} & \textbf{5.10 $\pm$ 0.89} & 7.37 $\pm$ 0.01  & \textbf{2.07 $\pm$ 0.00} \\
\bottomrule
\end{tabular}
\end{center}
\vspace{-0.2in}
\end{table*}

\begin{figure}[t]
    \begin{center}
    \centerline{\includegraphics[width=\columnwidth]{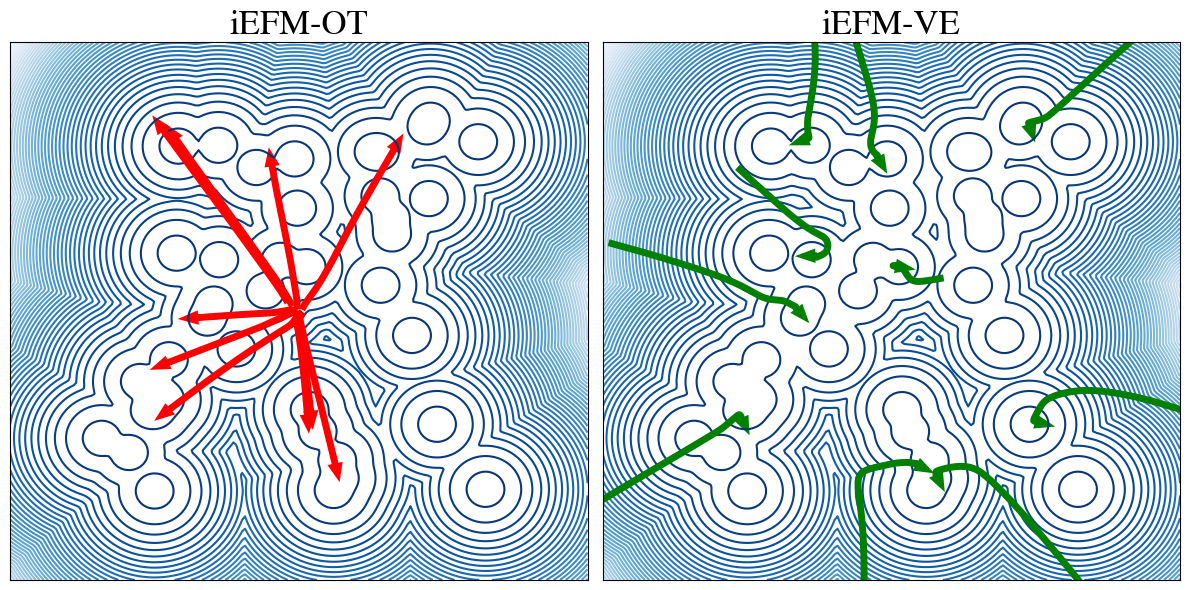}}
    \caption{Example of 10 trajectories on contour plot for the energy function of GMM. Colored line represent the trajectory of sampled ODE solution. Arrows represent the progression of time.}
    \label{fig:gmm_trajectory}
    \end{center}
    \vskip -0.4in
\end{figure}

We evaluate iEFM on two systems: a two-dimensional 40 Gaussian mixture model (GMM) and an eight-dimensional four-particle double-well potential (DW-4) energy function. For each energy function, we report negative log-likelihood~(NLL) and 2-Wasserstein distance~($\mathcal{W}_2$).

\textbf{Baselines.}
We compare iEFM to four recent works: the path integral sampler (PIS) \citep{zhang2021path}, denoising diffusion sampler (DDS) \citep{vargas2022denoising}, flow annealed bootstrapping (FAB) \citep{midgley2023flow}, and iterated denoising energy matching (iDEM) \citep{akhound2024iterated}. For baselines, we report the results from \citet{akhound2024iterated}.

\textbf{Architecture.} We parameterize the vector field $u_{\theta}$ using an MLP with sinusoidal positional embeddings for GMM and an EGNN flow model \citep{satorras2021n} for DW-4 following the prior work \citep{akhound2024iterated}.

\textbf{Metrics.} 
We use NLL and $\mathcal{W}_2$ as our metrics. For metric evaluation, we adopt pre-generated ground truth samples as datasets. Since we can generate ground truth samples in GMM, we adopt it as a dataset. In DW-4, we use a sample generated by MCMC in \citet{klein2024equivariant} as a dataset. Even though this is not a perfect ground truth, we believe it is a reasonable approximation of the ground truth samples.

To test NLL, we evaluate the log probability of the sample using a trained model. Except for iEFM, the probability is calculated using the optimal transport conditional flow matching~(OT-CFM) model \citep{tong2024improving}, which is trained in a sample-based training manner using a sample generated by each baseline method. For iEFM, NLL is directly evaluated by solving  ODE. To evaluate $\mathcal{W}_2$, we generate a sample from the trained model and measure $\mathcal{W}_2$ between generated sample and the dataset.

\textbf{Main results.} 
We report NLL and $\mathcal{W}_2$ for GMM and DW-4 in \cref{table:main table}. For GMM, iEFM matches or outperforms all considered baselines on NLL and $\mathcal{W}_2$. For DW-4, iEFM-OT outperforms all considered baselines on $\mathcal{W}_2$. 

In \cref{fig:gmm_sample}, we visualize the sample from iEFM of both probability paths with the sample from iDEM and ground truth GMMs. Notably, iEFM can capture the sharpness of GMM mode while the iDEM often samples slightly far points from the mode. This explains why iDEM significantly reduces $\mathcal{W}_2$ in GMM. Additionally, we visualize the trajectory of the ODE solution from both probability paths in \cref{fig:gmm_trajectory}. As expected, iEFM-OT has a straighter flow than iEFM-VE, even though there is no theoretical guarantee that the flow of iEFM-OT is marginally OT. 
\section{Conclusion}
In this work, we address the problem of sampling from Boltzmann distributions when only energy functions or unnormalized densities are available. To tackle this problem, we introduce iEFM, the first off-policy approach to training a CNF model. The EFM objective, a novel simulation-free off-policy objective that involves MC estimation of MVFs, enables iEFM to effectively utilize previously used samples with the replay buffer. Our experimental results demonstrate that iEFM either outperforms or matches iDEM on GMM and DW-4 benchmarks. Future research directions include validating and extending iEFM to complex high-dimensional systems. 


\bibliography{references}

\begin{thebibliography}{17}
\providecommand{\natexlab}[1]{#1}
\providecommand{\url}[1]{\texttt{#1}}
\expandafter\ifx\csname urlstyle\endcsname\relax
  \providecommand{\doi}[1]{doi: #1}\else
  \providecommand{\doi}{doi: \begingroup \urlstyle{rm}\Url}\fi

\bibitem[Akhound-Sadegh et~al.(2024)Akhound-Sadegh, Rector-Brooks, Bose, Mittal, Lemos, Liu, Sendera, Ravanbakhsh, Gidel, Bengio, et~al.]{akhound2024iterated}
Akhound-Sadegh, T., Rector-Brooks, J., Bose, A.~J., Mittal, S., Lemos, P., Liu, C.-H., Sendera, M., Ravanbakhsh, S., Gidel, G., Bengio, Y., et~al.
\newblock Iterated denoising energy matching for sampling from boltzmann densities.
\newblock \emph{arXiv preprint arXiv:2402.06121}, 2024.

\bibitem[Bengio et~al.(2021)Bengio, Jain, Korablyov, Precup, and Bengio]{bengio2021flow}
Bengio, E., Jain, M., Korablyov, M., Precup, D., and Bengio, Y.
\newblock Flow network based generative models for non-iterative diverse candidate generation.
\newblock \emph{Advances in Neural Information Processing Systems}, 34:\penalty0 27381--27394, 2021.

\bibitem[Berner et~al.(2024)Berner, Richter, and Ullrich]{berner2024an}
Berner, J., Richter, L., and Ullrich, K.
\newblock An optimal control perspective on diffusion-based generative modeling.
\newblock \emph{Transactions on Machine Learning Research}, 2024.
\newblock ISSN 2835-8856.
\newblock URL \url{https://openreview.net/forum?id=oYIjw37pTP}.

\bibitem[Chen et~al.(2018)Chen, Rubanova, Bettencourt, and Duvenaud]{chen2018neural}
Chen, R.~T., Rubanova, Y., Bettencourt, J., and Duvenaud, D.~K.
\newblock Neural ordinary differential equations.
\newblock \emph{Advances in neural information processing systems}, 31, 2018.

\bibitem[Del~Moral et~al.(2006)Del~Moral, Doucet, and Jasra]{del2006sequential}
Del~Moral, P., Doucet, A., and Jasra, A.
\newblock Sequential monte carlo samplers.
\newblock \emph{Journal of the Royal Statistical Society Series B: Statistical Methodology}, 68\penalty0 (3):\penalty0 411--436, 2006.

\bibitem[Klein et~al.(2024)Klein, Kr{\"a}mer, and No{\'e}]{klein2024equivariant}
Klein, L., Kr{\"a}mer, A., and No{\'e}, F.
\newblock Equivariant flow matching.
\newblock \emph{Advances in Neural Information Processing Systems}, 36, 2024.

\bibitem[Lahlou et~al.(2023)Lahlou, Deleu, Lemos, Zhang, Volokhova, Hern{\'a}ndez-Garc{\i}a, Ezzine, Bengio, and Malkin]{lahlou2023theory}
Lahlou, S., Deleu, T., Lemos, P., Zhang, D., Volokhova, A., Hern{\'a}ndez-Garc{\i}a, A., Ezzine, L.~N., Bengio, Y., and Malkin, N.
\newblock A theory of continuous generative flow networks.
\newblock In \emph{International Conference on Machine Learning}, pp.\  18269--18300. PMLR, 2023.

\bibitem[Lipman et~al.(2022)Lipman, Chen, Ben-Hamu, Nickel, and Le]{lipman2022flow}
Lipman, Y., Chen, R.~T., Ben-Hamu, H., Nickel, M., and Le, M.
\newblock Flow matching for generative modeling.
\newblock In \emph{The Eleventh International Conference on Learning Representations}, 2022.

\bibitem[Midgley et~al.(2023)Midgley, Stimper, Simm, Sch{\"o}lkopf, and Hern{\'a}ndez-Lobato]{midgley2023flow}
Midgley, L.~I., Stimper, V., Simm, G. N.~C., Sch{\"o}lkopf, B., and Hern{\'a}ndez-Lobato, J.~M.
\newblock Flow annealed importance sampling bootstrap.
\newblock In \emph{The Eleventh International Conference on Learning Representations}, 2023.
\newblock URL \url{https://openreview.net/forum?id=XCTVFJwS9LJ}.

\bibitem[Neal(2001)]{neal2001annealed}
Neal, R.~M.
\newblock Annealed importance sampling.
\newblock \emph{Statistics and Computing}, 11\penalty0 (2):\penalty0 125--139, 2001.

\bibitem[No{\'e} et~al.(2019)No{\'e}, Olsson, K{\"o}hler, and Wu]{noe2019boltzmann}
No{\'e}, F., Olsson, S., K{\"o}hler, J., and Wu, H.
\newblock Boltzmann generators: Sampling equilibrium states of many-body systems with deep learning.
\newblock \emph{Science}, 365\penalty0 (6457):\penalty0 eaaw1147, 2019.

\bibitem[Satorras et~al.(2021)Satorras, Hoogeboom, and Welling]{satorras2021n}
Satorras, V.~G., Hoogeboom, E., and Welling, M.
\newblock E (n) equivariant graph neural networks.
\newblock In \emph{International conference on machine learning}, pp.\  9323--9332. PMLR, 2021.

\bibitem[Sendera et~al.(2024)Sendera, Kim, Mittal, Lemos, Scimeca, Rector-Brooks, Adam, Bengio, and Malkin]{sendera2024diffusion}
Sendera, M., Kim, M., Mittal, S., Lemos, P., Scimeca, L., Rector-Brooks, J., Adam, A., Bengio, Y., and Malkin, N.
\newblock On diffusion models for amortized inference: Benchmarking and improving stochastic control and sampling.
\newblock \emph{arXiv preprint arXiv:2402.05098}, 2024.

\bibitem[Tong et~al.(2024)Tong, FATRAS, Malkin, Huguet, Zhang, Rector-Brooks, Wolf, and Bengio]{tong2024improving}
Tong, A., FATRAS, K., Malkin, N., Huguet, G., Zhang, Y., Rector-Brooks, J., Wolf, G., and Bengio, Y.
\newblock Improving and generalizing flow-based generative models with minibatch optimal transport.
\newblock \emph{Transactions on Machine Learning Research}, 2024.
\newblock ISSN 2835-8856.
\newblock URL \url{https://openreview.net/forum?id=CD9Snc73AW}.
\newblock Expert Certification.

\bibitem[Vargas et~al.(2022)Vargas, Grathwohl, and Doucet]{vargas2022denoising}
Vargas, F., Grathwohl, W.~S., and Doucet, A.
\newblock Denoising diffusion samplers.
\newblock In \emph{The Eleventh International Conference on Learning Representations}, 2022.

\bibitem[Wirnsberger et~al.(2022)Wirnsberger, Papamakarios, Ibarz, Racaniere, Ballard, Pritzel, and Blundell]{wirnsberger2022normalizing}
Wirnsberger, P., Papamakarios, G., Ibarz, B., Racaniere, S., Ballard, A.~J., Pritzel, A., and Blundell, C.
\newblock Normalizing flows for atomic solids.
\newblock \emph{Machine Learning: Science and Technology}, 3\penalty0 (2):\penalty0 025009, 2022.

\bibitem[Zhang \& Chen(2021)Zhang and Chen]{zhang2021path}
Zhang, Q. and Chen, Y.
\newblock Path integral sampler: A stochastic control approach for sampling.
\newblock In \emph{International Conference on Learning Representations}, 2021.

\end{thebibliography}
\bibliographystyle{icml2024}

\newpage
\appendix
\onecolumn
\section{Preliminary on Continuous Normalizing Flow}\label{appx:cnf}

Let $\mathbb{R}^d$ denote the data space and $\mu: \mathbb{R}^d \rightarrow \mathbb{R}_{>0}$ denote the density function of data-generating distribution. Continuous Normalizing Flow~(CNF) \citep{chen2018neural} is a deep generative model utilizing neural ordinary differential equation~(ODE) to describe $\mu$. In this section, we review the formal definition of CNF.

The CNF is defined by a time-dependent vector field $v_t: \mathbb{R}^d \rightarrow \mathbb{R}^d$ for $t \in [0, 1]$ which induces a flow map $\phi_t: \mathbb{R}^d \rightarrow \mathbb{R}^d$ via the following ODE:
\begin{equation}
\frac{d}{dt}\phi_t(x) = v_t(\phi_t(x)), \qquad \phi_0(x) = x
\end{equation}

The flow map $\phi_t(x)$ describes how point $x$ at time $t=0$ moves at given time $t$. With some well-behaved conditions on a vector field $v_t$, $\phi_t(x)$ is well-defined by the existence and uniqueness of the ODE solution of the initial value problem. Then, we can define the probability density path $p_t: \mathbb{R}^d \rightarrow \mathbb{R}^{+}$ for $t \in [0, 1]$ with simple given prior density $p_0$ as follows:
\begin{equation}
    p_t = [\phi_t]_* p_0
\end{equation}
If one can find a proper flow map $\phi_t$ which makes the distribution $p_1$ match with the target density $\mu$, sampling from the target density $\mu$ can be done by solving an ODE, based on the fact that $x_0 \sim p_0$ implies $\phi_1(x_0) \sim x_1$. In such case, the vector field $v_t$ is said to generate a probability density $p_t$ and target density $\mu$.

Now, we parameterize $v_t(x)$ with a neural network $u_t(x; \theta)$. The flow induced by $u_t(x, \theta)$ is called a Continuous Normalizing Flow~(CNF). Our goal is to find a proper $u_t$ so that its induced flow makes $p_1$ match with $\mu$. One way to achieve this is to make $u_t$ regress to $v_t$ that generates $\mu$ \citep{lipman2022flow}. This idea leads flow matching objectives in \cref{eq:fm}.


\end{document}